\pdfoutput=1
\documentclass[11pt]{article}

\usepackage[]{acl}

\usepackage{times}
\usepackage{latexsym}

\usepackage[T1]{fontenc}
\usepackage[utf8]{inputenc}

\usepackage{microtype}
\usepackage{times}
\usepackage{latexsym}
\usepackage{amsmath}
\usepackage{bbm}
\usepackage{graphicx}

\title{Differentiable Entailment for Parameter Efficient Few Shot Learning}

\author{Anonymous}
\author{Ethan Kim \\
 Harvard University \\
 \And
  Jerry Yang \\
    Harvard University \\}
    
\begin{document}

\maketitle

\begin{abstract}
Few-shot learning allows pre-trained language models to adapt to downstream tasks while using a limited number of training examples. However, practical applications are limited when all model parameters must be optimized. In this work we apply a new technique for parameter efficient few shot learning while adopting a strict definition of parameter efficiency. Our training method combines 1) intermediate training by reformulating natural language tasks as entailment tasks \cite{wang_entailment_2021} and 2) differentiable optimization of template and label tokens \cite{zhang_differentiable_2021}. We quantify the tradeoff between parameter efficiency and performance in the few shot regime and propose a simple model agnostic approach that can be extended to any task By achieving competitive performance while only optimizing 3\% of a model's parameters and allowing for batched inference, we allow for more efficient practical deployment of models.
\end{abstract}

\section{Introduction}
Large pre-trained language models have demonstrated adaptability to solve natural language processing (NLP) tasks. Typically, such language models are adapted to a downstream task through fine-tuning \cite{howard_universal_2018}. Although fine-tuning improves performance on downstream tasks, it is costly because it relies on updating every parameter of the model (355 million in the case of roBERTa)and requires storing a separate copy of the model for every downstream task. These storage requirements can become prohibitive, thus necessitating research into more parameter efficient methods . Alternative fine-tuning methods that update fewer parameters can have other tradeoffs. For example, adapter tuning fine-tunes a small number adapter parameters inserted between the transformer layers \cite{houlsby2019parameterefficient}but requires optimizing external parameters and still fine-tunes on the entire training dataset. Other methods have explored fine-tuning in the few shot learning case, where a limited number of labeled training samples are used for fine-tuning. These approaches have the disadvantages of relying on extreme model size \cite{DBLP:journals/corr/abs-2005-14165} \cite{lester_power_2021}, optimizing all model parameters \cite{wang_entailment_2021},\cite{zhang_differentiable_2021}, or using external architectures \cite{houlsby2019parameterefficient} \cite{li_prefix-tuning_2021} \cite{gao_making_2021}.
In this project, we present a simple, extensible method that improves few-shot performance without any extra parameters by combining two approaches: 1) leveraging trainable prompt pseudotokens rather than updating all the model parameters \cite{zhang_differentiable_2021}, and 2) reformulating  natural language processing tasks as entailment tasks and applying an intermediate training step, enabling better generalization to downstream tasks. \cite{wang_entailment_2021}. Our major contributions are as follows. 

\begin{itemize}
    \item Our method achieves competitive few shot performance while optimizing only 3\% of a model's parameters reducing storage costs by a factor of 30. 
    \item We introduce a strict definition of parameter efficiency which extends the practical uses of few shot learning by allowing batching of computation across tasks.
\end{itemize}

\section{Related Work}
\subsection{Finetuning}
The standard method for fine-tuning Masked Language Models (MLMs) like BERT applies a classification head to the [CLS] token representation. The language model learns to update the [CLS] representation to better solve the downstream task. A number of reformulations have been proposed seeking to increase performance and improve parameter efficiency. 

\subsection{Prompting}
Language models learn a general set of abilities that can be adapted to specific downstream tasks. One method is to use task-specific natural language prompts to guide the language model output. GPT-3, for example, uses prompts and in-context examples to achieve good few-shot performance on various tasks \cite{DBLP:journals/corr/abs-2005-14165}. GPT-3 leverages extreme scale (175 Billion parameters) to adapt to natural language prompts without fine-tuning. Prompting can be particularly useful for few-shot learning in the low-data regime. For some tasks, a well designed prompt can be shown to be equivalent to hundreds or thousands of additional labeled training points \cite{le-scao-rush-2021-many}. AUTOPROMPT uses a gradient-based search to optimize a discrete prompt \cite{DBLP:journals/corr/abs-2010-15980}. LMBFF uses an auxiliary language model to generate a set of candidate prompts and chooses the best candidate \cite{gao_making_2021}.

\subsection{Pattern Exploiting Training}
One alternative to standard fine-tuning is to model the output as a cloze completion task where the output is the model's representation of a masked input token \cite{schick2021exploiting}. Intuitively, this approach works well because it more closely matches the training process for MLMs. In the pre-training task for models such as BERT and roBERTA, the model is asked to predict the identity of a masked token based on the hidden representations of neighboring tokens.

Further work has extended this approach to use natural language prompts to guide the cloze output \cite{gao_making_2021}. Additional work has focused on training the prompt tokens in continuous space by optimizing a set of prompt pseudotokens. \cite{li_prefix-tuning_2021}
\cite{liu_gpt_2021}
\cite{lester_power_2021}. Additionally in the DART method, the tokens used as classification labels can be optimized  \cite{zhang_differentiable_2021}.

\subsection{Entailment Reformulation}
Work from \cite{wang_entailment_2021} focuses on improving language model performance by formulating NLP tasks as entailment tasks. Fundamentally, entailment seeks to determine whether for a pair of inputs $(S_1, S_2)$, the first sentence entails or contradicts the second. Most standard  classification tasks in NLP can be reformulated as entailment tasks. For example, a sentiment analysis task can be be framed as an entailment task using the following template: 
\begin{align}
    [\text{CLS}] S_1[\text{SEP}] S_2 [\text{EOS}],
\end{align}
With $S_2$ = "It was great" as the entailment prompt. Instead of using the [CLS] token representation of $S_1$ to classify the review as positive or negative as in standard fine-tuning, we instead concatenate the text with the prompt and use the [CLS] token representation of the concatenated sequence to denote whether the first sentence entails the second. 

For multi-class classification problems we construct a different input for every class and take the label as the class with the highest entailment score. 

A key to the success of the entailment approach from \cite{wang_entailment_2021} is an intermediate training step where the pre-trained language model is fine-tuned on a natural language inference (NLI) task like MNLI. Intuitively, the model can be adapted to be good at one entailment task and then generalize to perform well on other reformulated entailment tasks. 

\subsection{Parameter Efficiency}

\begin{figure*}
    \centering
    \includegraphics[width=12cm, height=6cm]{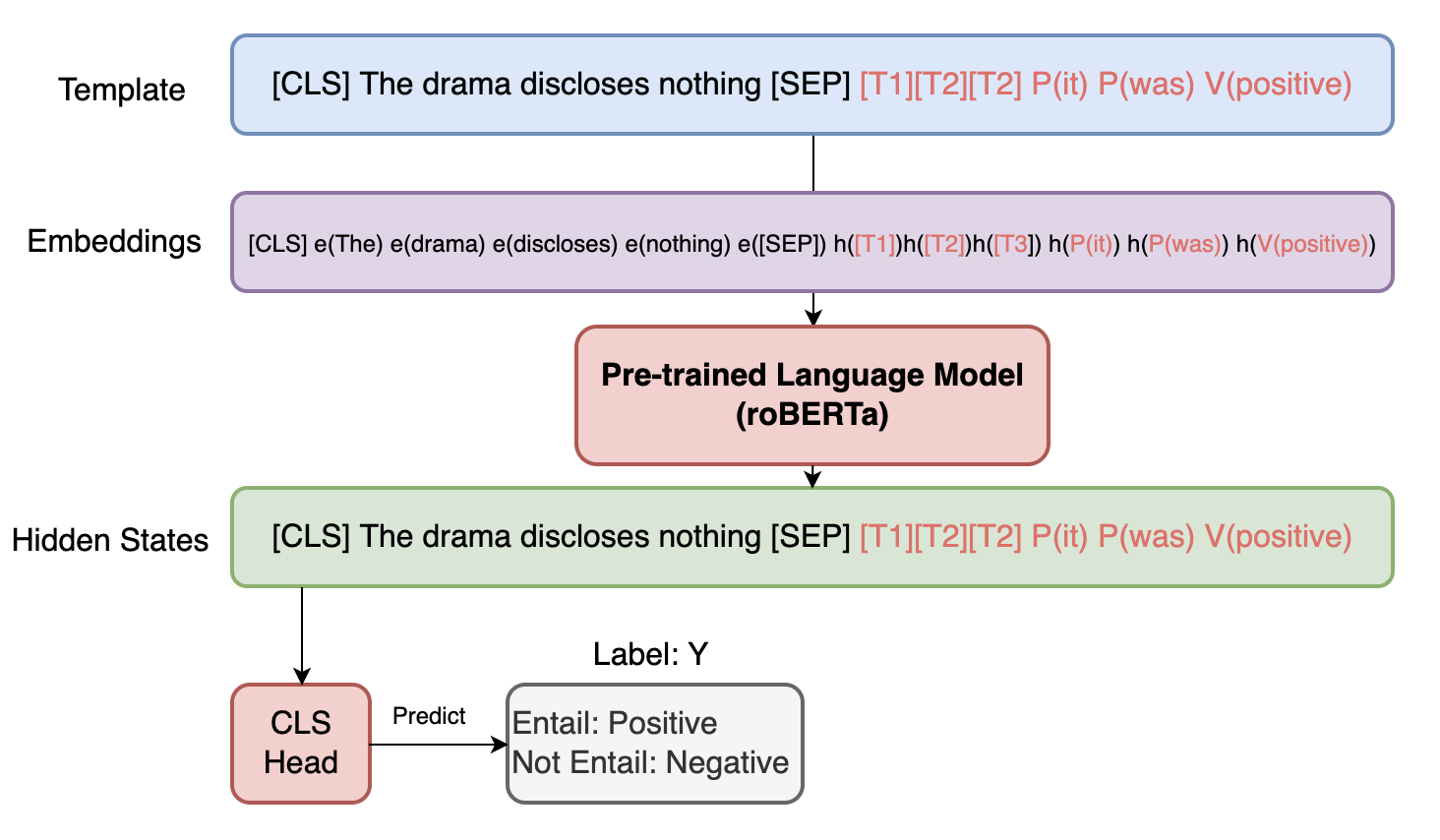}
    \caption{Differential Entailment Approach}
    \label{fig:my_label}
\end{figure*}

Related works adopt various, sometimes contradictory, definitions of parameter efficiency when applied to language model fine-tuning. Broadly, these definitions can be grouped into several
categories: 
\begin{enumerate}
    \item reducing the number of model parameters necessary to achieve good few shot adaptability
    \item optimizing a small subset of the total model parameters
    \item avoiding external parameters or changes to the model architecture
\end{enumerate}

Some works on few-shot learning explore techniques allowing smaller models to learn robustly \cite{wang_entailment_2021}.
Large models such as GPT3 with 175 billion parameters can take advantage of their scale to perform well at few-shot in-context learning \cite{DBLP:journals/corr/abs-2005-14165}. A technique can be parameter efficient if it allows similar results be be achieved with a smaller language model e.g. a  340 million parameter roBERTa model rather than the 175 billion parameter GPT-3 or 11 billion parameter T5.

Parameter efficiency can also aim to optimize a smaller number of task specific parameters while keeping most of the language model parameters frozen. Adapter tuning inserts trainable layers between the frozen layers of a Transformer language model. \cite{houlsby2019parameterefficient}. Prompt tuning optimizes a small set of trainable input tokens while keeping the pre-trained Transformer layers frozen \cite{lester_power_2021}. Lite Self Training (LiST) freezes most of the encoder parameters and only trains a small number of adapter parameters \cite{wang2021list}. LoRA, adds low rank trainable matrices between transformer layers \cite{DBLP:journals/corr/abs-2106-09685} while freezing the pretrained model.

Other works define parameter efficiency as the lack of a need for parameters external to the model being fine-tuned. Part of the motivation for differential prompt tuning \cite{zhang_differentiable_2021} is that it directly optimizes trainable pseudotokens without the need for an external model such as LSTM in P-tuning \cite{liu_gpt_2021}. Such approaches are advantageous as they require no modifications to a pre-trained model's architecture and do not add additional inference time like adapters. Delta Tuning explores in depth the performance of different parameter efficient approaches at different model scales and in combination with one another \cite{ding2022delta}

We focus on parameter efficiency in the true few shot learning regime. Therefore, we do not take advantage of any additional unlabeled training data. Iterative PET used this approach to pseudolabel unlabelled samples and provide extra training examples to a model \cite{schick2021exploiting}. LiST iteratively trains a student model on data pseudolabeled by a teacher model \cite{wang2021list}. However, these semi-supervised learning approaches require extra unlabeled training data as well as additional training computation compared to true few-shot learning.

\section{Approach}

Our main approach is shown in Figure \ref{fig:my_label}. We convert all NLP tasks to the entailment format and train few shot models from an intermediate training checkpoint. The entailment approach outlined in \cite{wang_entailment_2021} performs traditional fine-tuning and updates all model parameters via gradient descent. Instead of performing the computationally expensive update step on all model parameters, our approach fine-tunes only the prompt and label tokens in continuous space while keeping the main language model frozen. By using more expressive pseudotokens as part of our prompt and by training only the input parameters, we achieve a parameter efficient few shot learning method with competitive few-shot performance. 

\subsection{Pseudotokens}

With discrete tokens, the label template tokens are either chosen manually or determined through a search over tokens in a discrete space. In comparison, our label descriptions are optimized in continuous space via back-propagation and hence can attain more expressive, fine-grained representations to prompt a model for a certain task. Formally, we define a
set of pseudotokens $\mathcal{T} \notin \mathcal{V}$ outside of the normal vocabulary. The pseudotoken embedding $h(\mathcal{T})$ is a trainable set of parameters that are optimized via backpropagation. For a given input we might have the following prompt:
$$
S_1 \text{[SEP] } \mathcal{T}_0 \mathcal{T}_1 \mathcal{T}_2 \text{ it was} \text{ [LABEL]}
$$
We differentiably optimize prompting pseudotokens. We also experiment with allowing the label embedding $h(\text{[LABEL]}$ to be a pseudotoken with a trainable embedding.
For label and prompt tokens we experiment with both initializing these pseudotokens embeddings randomly and initializing them with the embeddings of the original tokens.

\begin{figure*}
\centering
\includegraphics[width=12cm, height=6cm]{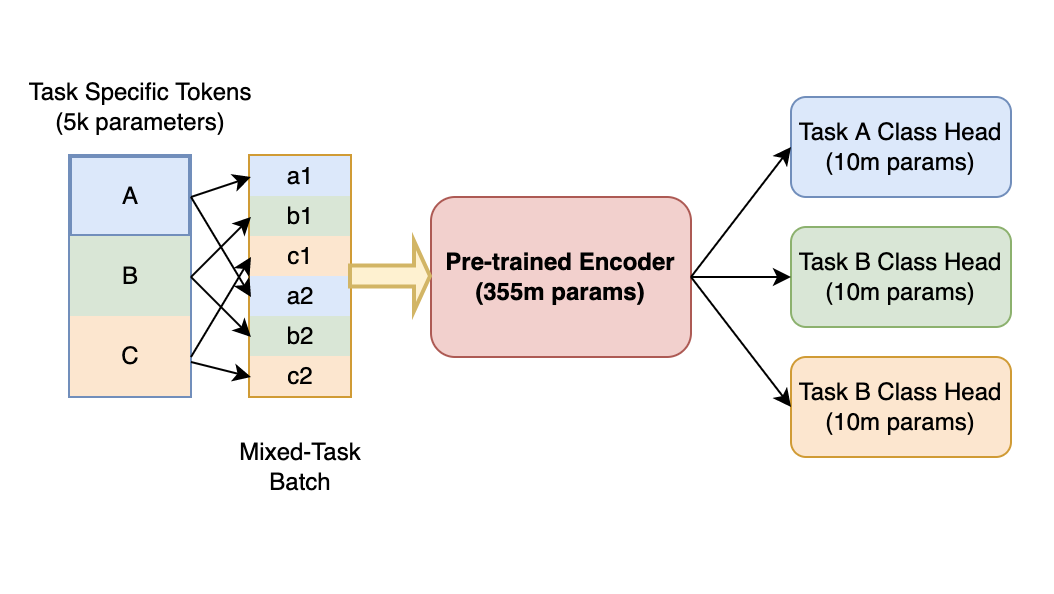}
\caption{Entailment allows batching of hidden state computations across tasks}
\end{figure*}
\subsection{Parameter Efficiency}

We adopt the strictest definition of parameter efficiency that has practical advantages for downstream applications. In Differentiable Entailment we 1) use a smaller language model compared to GPT-3 or T5, 2) freeze the main encoder parameters, 3) only fine-tune a limited set of pseudotokens without any external parameters or architectural modifications and 4) employ strict few-shot learning without using any additional training data. 

Following the method in Prompt Tuning, we freeze the main model parameters and only fine-tune the subset of trainable input tokens \cite{lester_power_2021}. In contrast to Prompt Tuning we also fine tune the model classification head since we are outputting a specific classification label rather than using a generative model such as T5.

By freezing the model parameters we can efficiently optimize a smaller set of task-specific parameters, namely the pseudotoken embeddings as well as the entailment classification head. In contrast to approaches outlined above, which rely on a large-scale model to make up for a reduction in trainable parameters \cite{lester_power_2021}, we use a smaller language model. With roBERTa-large this leads to a more than 30x reduction in the number of trainable parameters. Furthermore, instead of storing a fine-tuned 355 million parameter model for each task, we only need to store the task-specific trainable pseudotoken embeddings and classification head. Finally, in contrast to methods which finetune all the model parameters \cite{zhang_differentiable_2021} \cite{wang_entailment_2021} or methods with external parameters \cite{houlsby2019parameterefficient} 
our method allows the hidden state computation for different tasks to be batched together since only the specific prompt embeddings for each tasks need to be changed. As others have noted: such \textbf{in batch parallel computing} has extreme practical application \cite{ding2022delta}. LoRA also allows for multitask batching, however applying additional low rank matrices to later transformer layers is more complex than simply swapping out a set of task specific input embeddings \cite{DBLP:journals/corr/abs-2106-09685}.

\subsection{Templates}

We explore several different approaches to combining label templates with pseudotokens. For various tasks, we adapt the standard prompt templates used in previous  works \cite{zhang_differentiable_2021} \cite{wang_entailment_2021}. For example, sentiment analysis tasks such as CR can be prompted for both entailment and cloze completion in a simple way. In \ref{table:template}, we show label templates for a sentiment analysis tasks. For such tasks, the prompt standard template is "it was great". The cloze completion method concatenates the prompt to the input sentence and masks out the label "great", whereas our method concatenates the template without masking the token of interest and predicts entailment. When training label templates in the continuous space, we initialize from the embeddings of the label template tokens in the standard template. For example, given the following template:
$$
S_1 \text{[SEP]}  \mathcal{T}_0 \dots \mathcal{T}_j\text{it was } \text {great}
$$
We would train the prompt tokens "it", "was", the label token "great" and $j + 1$ additional pseudotokens.

For sentence pair tasks such as Quora Question Pairs (QQP), we adopt a slightly different template following \cite{wang_entailment_2021}\cite{zhang_differentiable_2021}. The task is to predict entailment based on the sentence pairs and a set of prompt pseudotokens inserted between them. For QQP we use the format 
$$
S_1 \text{[SEP]} \mathcal{T}_0 \dots \mathcal{T}_j S_2
$$ 

\begin{table}[t]
\centering
\resizebox{\columnwidth}{!}{%
\begin{tabular}{ll}
\hline
Method                        & Template                                            \\ \hline
Cloze                         & S\_1 {[}SEP{]} it was {[}MASK{]}                    \\
Entailment                    & S\_1 {[}SEP{]} it was great                         \\
Differential Prompt           & S\_1 {[}SEP{]}{[}Prompt tokens{]} great             \\
Differential Label and Prompt & S\_1 {[}SEP{]}{[}Prompt tokens{]} {[}Label token{]} \\ \hline
\end{tabular}%
}
\caption{Example Prompting Templates for a Sentiment Classification task. For our method we optimize either a set of prompt pseudotokens and/or a label pseudotoken.}
\label{table:template}
\end{table}

\subsection{Symmetry of Entailment}

In \cite{wang_entailment_2021}, a single label description $p$ is used for each example in a binary classification task, e.g. a binary sentiment classification task is formulated as whether input sentence $S_1$ entails $S_2 = \text{"This indicates positive sentiment."}$. To encourage more robust tuning of the label description parameters and classification head, we experiment with using two label descriptions $p_1$ and $p_{-1}$ for binary classification tasks, and augment the dataset as:
\begin{align}
    \mathcal{D}_{\text{train}} = \{(x_i, p_{1}, y_i) \cup (x_i, p_{-1}, -y_{i})\}_{i=1}^{K}
\end{align}
For a positive sentiment example, the two corresponding samples in the training dataset would be $(x_i, p_{1}, 1)$ and $(x_i, p_{-1}, -y_{i})$ where $p_{1} = \text{This indicates positive sentiment}$ with label 1 (does entail) and $p_{-1} = \text{This indicates negative sentiment}$ with label 0 (does not entail).

\begin{figure}
    \centering
    \includegraphics[width=8cm, height=4cm]{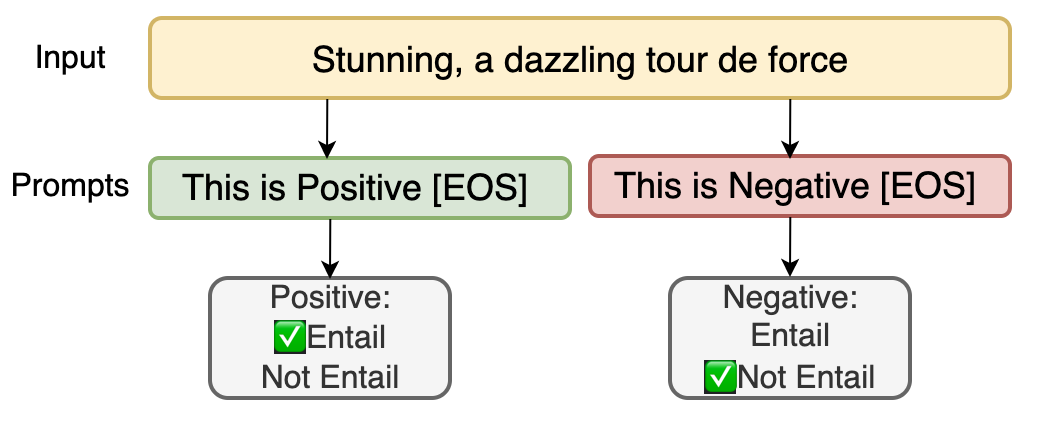}
    \caption{Symmetry for simple data augmentation}
    \label{fig:sym}
\end{figure}

\section{Experiments}

\begin{table*}
\centering
\begin{tabular}{lllllll}
\hline
   & SST-2          & MR         & CR         & MPQA       & Subj       & CoLa       \\ \hline
\multicolumn{2}{l}{Full   Training Dataset} &            &            &            &            &            \\ \hline
Majority                   & 50.9           & 50         & 50         & 50         & 50         & 69.1       \\
Finetuning                 & 95             & 90.8       & 89.4       & 89.4       & 97         & 86.2 (1.6) \\
EFL                        & 96.9 (0.2)     & 92.5 (0.1) & 92.5 (0.4) & 90.8 (0.4) & 97.1 (0.2) & 86.4 (0.5) \\ \hline
\multicolumn{2}{l}{Few Shot k =   16}       &            &            &            &            &            \\ \hline
Fine Tuning                & 81.4 (3.8)     & 76.9( 5.9) & 75.8 (3.2) & 59.0 (3.4) & 90.8 (1.8) & 70.0 (0.9) \\
DARTS                      & 93.5 (0.5)     & 88.2 (1.0) & 91.8 (0.5) & 85.6 (0.3) & 90.7 (1.4) & -          \\
LMBFF                      & 92.3 (1.0)     & 85.5 (2.8) & 91.0 (0.9) & 85.8 (1.9) & 91.2 (1.1) & 69.5 (0.5) \\
EFL                        & 90.8 (1.0)     & 86.2 (0.8) & 92.3 (0.4) & 87.0 (0.6) & 80.0 (5.4) & 69.4 (0.9) \\
DE                         & 91.9 (0.5)     & 87.1 (2.1) & 91.5 (1.4) & 87.0 (0.9) & 89.5 (2.4) & 70.3 (2.4) \\
DE PE                      & 91.1 (0.2)     & 84.5 (0.3) & 91.6 (0.2) & 85.9 (0.6) & 81.5(0.1)  & 69.7 (0.3) \\ \hline
\end{tabular}

\caption{Main Results: all results use roBERTa-large as the base architecture, the standard deviation across 5 training folds is given. Differentiable Entailment (DE) is our method fine-tuning all model parameters. Differentiable Entailment Parameter Efficient (DE PE) is our parameter efficient method which only finetunes the trainable pseudotokens and classification head.}
\label{table:main}
\end{table*}
\subsection{Evaluation}
We evaluate our method on the tasks from \cite{wang_entailment_2021} which are mainly the subset of the GLUE and SuperGLUE benchmark tasks that are compatible with the entailment reformulation. In addition, we  follow the best practices for evaluation of few shot NLP fine-tuning methods \cite{bragg_flex_2021}. For each experiment we sample 5 non-overlapping training folds and report average performance after k-shot training over the entire test set \cite{gao_making_2021}. Hyperparameters are tuned for each task and method.

\subsection{Implementation Details}
Models are implemented using the pytorch \cite{DBLP:journals/corr/abs-1912-01703} and transformers \cite{DBLP:journals/corr/abs-1910-03771} libraries with code adapted from \cite{zhang_differentiable_2021}. Our pre-trained model is roBERTa large \cite{DBLP:journals/corr/abs-1907-11692}. Checkpoints for roberta-large-base as well as checkpoint models are downloaded from huggingface. We experiment with different intermediate checkpoints, namely $\text{roberta-large-mlni}$ and a checkpoint trained robustly on a wide variety of NLI tasks (adversarial NLI /ANLI)\cite{nie-etal-2020-adversarial}. Experiments were run using approximately 100 GPU hours on a single V100. 

\subsection{Results}

Table \ref{table:main} contains main results for single sentence classification   tasks. Table \ref{table:pair} shows results for various sentence pair tasks. 
We compare our approach with other few shot learning techniques and experiment with various modifications to the differential entailment method.

\begin{table}
\centering

\begin{tabular}{lll}
\hline
                   & MRPC             & QQP        \\ \hline
                   
        Full Training Dataset           & (f1)             & (f1)       \\
        \hline
Majority           & 81.2             & 0          \\
Finetuning         & 89.9 (1.7)       & 89.0 (0.1) \\
EFL                & 91.0 (0.8)       & 89.2 (0.1) \\\hline
\multicolumn{2}{l}{Few Shot k =   16} &            \\\hline
Fine Tuning        & 76.6 (2.5)       & 60.7 (4.3) \\
DARTS              & 78.3 (4.5)       & 67.8 (3.2) \\
LMBFF              & 76.2 (2.3)       & 67.0 (3.0) \\
EFL                & 76.2 (1.3)       & 67.3 (2.6) \\
DE                 & \textbf{83.3} (0.1)       & \textbf{72.9} (0.3) \\
DE PE              & 78.0 (1.5)       & 72.6 (0.7) \\ \hline
\end{tabular}

\caption{Results for sentence pair tasks. NLI tasks such as MNLI, QNLI and SNLI are excluded from the comparison because these datasets are already incorporated as part of the intermediate training step for the ANLI model}
\label{table:pair}
\end{table}

\begin{table}
\begin{tabular}{lll}
\hline
Tokens & SST2       &  \\
\hline
0      & 90.5 (0.4) &  \\
2      & \textbf{91.1} (0.7) &  \\
5      & \textbf{91.1} (0.2) &  \\
20     & 90.6 (0.5) & \\ \hline
\end{tabular}

\caption{Performance Scaling with number of trainable pseudotokens. Using a set of 5 trainable  pseudotokens performed best.}
\label{table:scale}
\end{table}

\subsection{Intermediate Training}

We experiment with different intermediate training steps.  Table \ref{table:intermediate} shows results for fine-tuning various checkpoints. The MNLI and ANLI checkpoints drastically outperform the roberta-base checkpoint because they have been adapted to perform well on entailment tasks. The ANLI model was trained on multiple augmented entailment tasks\cite{wang2021list} and offers a further boost in performance. These results show that the entailment reformulation relies heavily fine-tuning a model that has already been adapted for entailment.

\begin{table}
\begin{tabular}{llll}
\hline
      & Base & MNLI       & ANLI       \\ \hline
SST-2 & 50.1 (0.1)         & 89.8 (1.3) & \textbf{91.1} (0.2) \\
MR    & 51.1 (0.2)         & 83.6 (0.4) & \textbf{84.5} (0.3) \\ \hline
\label{table:intermediate}
\end{tabular}

\caption{Importance of Intermediate Training Steps: Accuracy is shown for finetuning from the roberta-large checkpoint, a checkpoint trained on MNLI, and a checkpoint trained on ANLI, a large number of curated and synthetic NLI examples. The more robustly trained NLI checkpoint consistently performs better on downstream tasks.}
\label{table:intermediate}
\end{table}

\subsection{Prompting Schemes}

We further experiment with different prompting schemes. We find best performance when we train the prompt tokens, the label token and an additional set of task specific pseudotokens. Table \ref{table:scale} shows scaling with various numbers of prompting pseudotokens. Using 5 additional pseudotokens in addition to trainable prompt and label tokens worked best.

\subsection{Symmetry}

By adding an symmetric entailment example for binary classification tasks during training we can effectively provide double the training signal (Figure \ref{fig:sym}). However, it appears that it is difficult for the model to learn from the two complementary training signals in a few shot scenario. Simply adding the symmetric examples at training time leads to a drop in performance (Table \ref{table:sym}). These results reveal limitations in the model's actual understanding of the entailment task. When given only the template with the positive label the model learns to associate entailment with the positive class and not entailment with the negative class. When using additional symmetric examples, this correlation is reversed and may be too difficult for a model of this size and ability to parse. Further work could explore improving this method or ensembling the outputs of models trained on symmetric examples.

\begin{table}
\begin{tabular}{llll}
\hline
       & SST-2      & MR         & CR         \\ \hline
DE PE     & \textbf{91.1} (0.2) & \textbf{84.5} (0.3) & \textbf{91.6} (0.2) \\
DE PE Sym & 51.1 (3.1) & 48.3 (3.3) & 52.2(2.4) \\ \hline
\end{tabular}
\caption{Few shot learning results on binary classification using symmetric entailment scheme. DE PE is the regular parameter efficient differential entailment method. Training with both symmetric signals does not lead to a robust model.}
\label{table:sym}
\end{table}

\section{Analysis and Discussion}

Our method achieves competitive performance with other few shot learning techniques while optimizing 30 times fewer parameters. On most single sentence tasks performance is within a few points of methods that train all model parameters. When we relax the constraints on parameter efficiency performance is directly competitive with other few shot learning methods. In some cases we exceed the performance of methods that rely on optimizing all model parameters or even additional external architectures. Notable we achieve much stronger performance on sentence pair tasks such as MRPC and QQP. We theorize that this may be because these sentence pair tasks are most similar to the entailment tasks seen during intermediate training.

Fundamentally, intermediate training is crucial for parameter efficient performance because it gives the model a head start in adapting to the reformulated task. We see that using a strong NLI trained intermediate model improves results (Table \ref{table:intermediate}). To adapt to a specific entailment task then requires only a small number of parameter updates.

\section{Conclusion}

In this paper we achieve parameter efficient few-shot learning by combining 1) entailment reformulation of NLP tasks and 2) trainable prompt pseudotokens in the continuous space. Our Differentiable Entailment approach achieves competitive results while only training 3\% of the parameters compared to match. We quantify the impact of intermediate training steps and different prompting schemes. By adopting a  strict definition of a parameter efficiency we achieve few-shot performance with fewer trainable parameters, no external parameters and without scaling up model size or using unlabeled training data. One major limitation is that we have to train a separate classification head for each downstream task, limiting potential gains in parameter efficiency. Further work could explore different intermediate training tasks, ensembling sets of prompts tokens and combining cloze completion for classification with the entailment reformulation. Given that our method is model agnostic and efficient it is likely to be broadly applicable to additional tasks.

\section{Broader Impact}
Parameter efficient models, especially with the method described in this paper have the potential to allow use of machine learning models on a more widespread basis. In our approach, batching computations for different tasks and using a single forward pass through a model could allow many models to be run on a single device at a single team. Such a scheme has advantages in terms of providing more accessibility to machine learning models and reduced energy consumption. However, parameter efficiency also opens that door to running personalized models that may be injurious to individual security or privacy. For example, user specific embeddings could easily be trained to predict a user's behavior with a specialized model. We anticipate that such potential use cases of parameter efficient few shot learning should be treated carefully.
\bibliography{anthology,custom}
\bibliographystyle{acl_natbib}

\appendix
\section{Hyperparameters}

The hyperparameter search space used for all experiments is as follows:
\begin{itemize}
    \item learning rate [1e-5, 3e-5, 1e-4]
    \item weight decay [0.0, 0.05, 0.1]
    \item batch size [8, 16]
    \item gradient accumulation steps [1, 2]
\end{itemize}

\section{Prompting Templates}
The standard prompting templates from \cite{wang_entailment_2021} are used for each task.
\begin{itemize}
    \item SST-2:  sentence1[SEP]It was great
    \item MR:  sentence1[SEP]It was great
    \item CR:  sentence1[SEP]It was great
    \item MPQA:  sentence1[SEP]It was positive
    \item Subj: sentence1[SEP]It was objective
    \item CoLA: sentence1[SEP]It was correct
    \item MRPC:  sentence1[SEP]sentence2
    \item QQP: sentence1[SEP]sentence2
\end{itemize}

\end{document}